\documentclass[letterpaper, 10 pt, conference]{ieeeconf}  

\IEEEoverridecommandlockouts                              

\usepackage{graphicx}
\usepackage{algorithm}
\usepackage{algorithmic}
\usepackage{amsmath}
\usepackage{amsfonts}
\usepackage{float}
\usepackage{booktabs}
\usepackage[labelformat=empty]{subfig}
\usepackage{cite}
\usepackage{amsmath,amssymb,amsfonts}
\usepackage{textcomp}
\usepackage{xcolor}
\usepackage{hhline}

\let\labelindent\relax
\usepackage[inline,shortlabels]{enumitem}

\title{\LARGE \bf
Driving behavior recognition via self-discovery learning
}

\author{Yilin Wang$^{1}$ 
}

\begin{document}

\maketitle
\thispagestyle{empty}
\pagestyle{empty}

\begin{abstract}

Autonomous driving systems require a deep understanding of human driving behaviors to achieve higher intelligence and safety. Despite advancements in deep learning, challenges such as long-tail distribution due to scarce samples and confusion from similar behaviors hinder effective driving behavior detection. Existing methods often fail to address sample confusion adequately, as datasets frequently contain ambiguous samples that obscure unique semantic information. To tackle these issues, we propose a Self-Discovery Learning (SDL) framework that captures subtle variations in driving behaviors through intrinsic pattern exploration and distinctively handles confusing samples. Our approach employs a transformer-based feature extractor to model long-range temporal dependencies and a sub-pixel level temporal discovery framework to enhance the network’s ability to discern finer spatiotemporal differences. Additionally, we introduce a sample discovery mechanism to dynamically update features and estimate sample uncertainty, enabling the network to focus on semantically correct samples. The proposed SDL framework is plug-and-play, introduces no additional parameters during inference, and demonstrates superior performance in extensive quantitative and qualitative experiments.

\end{abstract}

\section{INTRODUCTION}

In recent years, autonomous driving has garnered significant attention. With the advancements in deep learning technologies, numerous tasks have been effectively addressed. However, a deeper comprehension of human driving behaviors and patterns is essential to realize a highly intelligent transportation system. This includes understanding how drivers interact with other participants in traffic scenarios and how they make decisions in various traffic conditions. Such an in-depth understanding of driving behaviors is critical for the development of more intelligent and safer autonomous driving systems\cite{kaplan2015driver}.

To address these challenges, researchers have compiled and developed comprehensive driving behavior datasets\cite{ramanishka2018toward}, proposing various research frameworks from multiple perspectives. The primary challenges in detecting driving behaviors currently include: 1) the long-tail distribution resulting from the scarcity of samples; and 2) the confusion arising from similar behaviors or manual annotations. Consequently, models tend to be misled by redundant information rather than focusing on distinctive driving behaviors, leading to suboptimal performance.
\begin{figure}[htbp]
  \centering
  \includegraphics[width=\columnwidth]{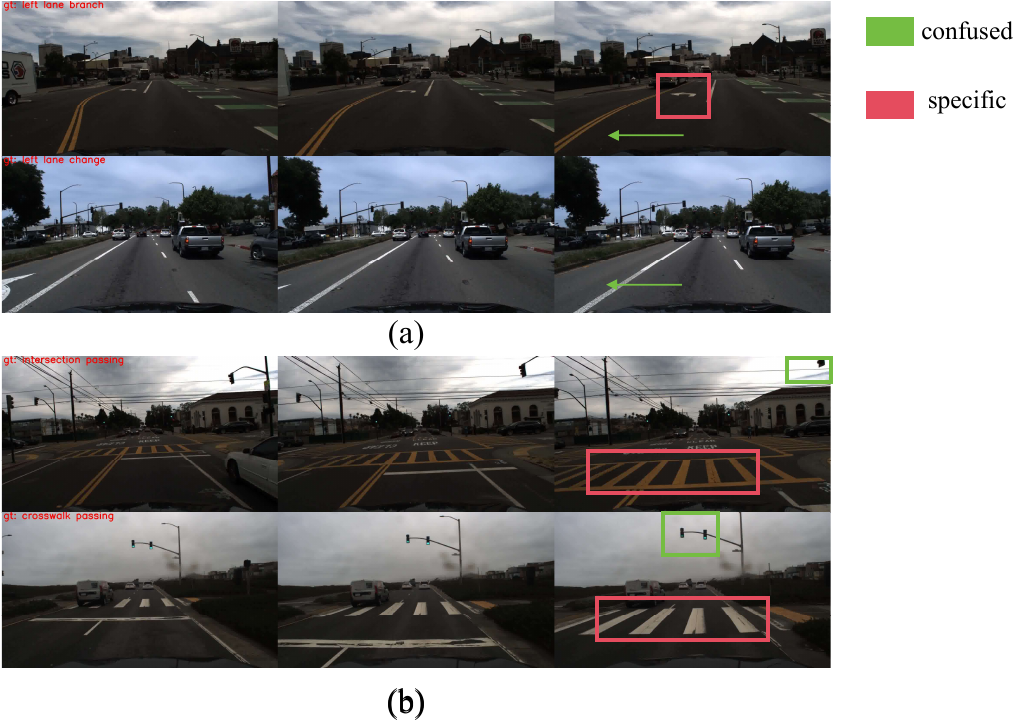}
  \caption{The confused samples in HDD dataset. Red denotes specific information, whereas green signifies confusing information. “Left branch change” indicates a vehicle exits the main road to take a left branch or fork, "Left lane change" indicates a vehicle exits the main road to take a left branch or fork, "Intersection passing" indicates driving through an intersection where two or more roads meet, "Crosswalk passing" indicates driving through a designated pedestrian crossing area.}
  \label{fig:intro}
\end{figure}

To mitigate long-tail issue, Noguchi et al.\cite{noguchi2023ego} has attempted to incorporate additional data to counteract the imbalance in scenarios with few samples. Meanwhile, cao et al.\cite{cao2023e2e} has designed a task-specific network architecture to extract more robust spatiotemporal features. However, these approaches do not adequately address the challenge of sample confusion. Existing datasets often contain ambiguous samples, making it difficult for models to learn unique semantic information. For instance, as illustrated in Figure.\ref{fig:intro}, the driving behaviors "left lane branch" and "left lane change" both involve left-turn actions. The distinction between them must be discerned from the "left-turn" lane markings in the background. Therefore, enabling models to capture finer spatiotemporal differences and distinguish unique characteristics between similar scenarios is paramount in the understanding of driving scenes.

In this work, we propose a Self Discovery Learning (SDL) framework to capture subtle variations in driving behavior through in-depth exploration of intrinsic patterns and to handle different confusing samples distinctively. Given that the differences in various driving behaviors are often subtle and not easily perceptible, we employ the spatial-temporal transformer\cite{dosovitskiy2020image} framework as a feature extractor to capture long-range dependencies across different temporal frames, thereby obtaining robust feature representations. Inspired by the strong capability of self-supervised tasks\cite{jing2020self} in modeling deep abstract semantic features, we construct a sub-pixel level reconstruction learning framework for input sequences. This aims to enable the network to capture finer temporal differences and more distinctive spatial information during the reconstruction learning process. 

To further distinguish similar scenarios, we first establish a set of dictionary vectors to represent different categories in high-dimensional space. By leveraging the metric relationship between the dictionary vectors and the extracted features, we dynamically update the features and the dictionary in real-time, aiming to make the feature space of different categories more distinct at the classification boundary. Additionally, utilizing the benchmark role of the dictionary vectors, we estimate the uncertainty of the input frame features as a measure of sample confusion. This uncertainty is then backpropagated through the loss function, allowing the network to learn and dynamically select semantically correct samples. It is important to note that this framework is effective only during training and does not introduce additional parameters during the inference phase.

In summary, our main contributions are as follows:

1) We propose a sub-pixel level temporal discovery framework to capture dynamic temporal and spatial semantic information hidden in the input sequence. Without introducing additional data labels, SDL encourages the network to learn hidden abstract semantics and temporal information through self-supervised learning between the original frames and the reconstructed frames.

2) To address the issue of semantic confusion, we introduce a sample discovery framework. By leveraging the benchmark role of the dictionary vectors, we dynamically update the dictionary and features and further estimate the uncertainty of the features. This serves as a measure of sample confusion, further aiding the network in dynamically selecting samples.

3) The proposed SDL is plug-and-play and can be applied to different framework architectures. More importantly, it does not introduce additional parameters during the inference phase.

4) we conduct extensive quantitative and qualitative experiments. The experimental results demonstrate the superiority and effectiveness of our proposed SDL.

\section{Related work}
\subsection{Driving Behavior Recognition}
Over the recent years, a number of works have been proposed for driving behavior recognition\cite{kuge2000driver, jain2015car, oliver2000graphical}. In the early stage, Hidden Markov Models (HMM) were widely used for modeling driving behaviors. For example, Jain et al.\cite{jain2015car} combined in-vehicle and external contextual information to predict driving maneuvers such as lane changes and turns several seconds in advance, demonstrating effectiveness on 1180 miles of naturalistic driving data. Oliver et al.\cite{oliver2000graphical} integrated graphical models with HMM to identify seven distinct driving maneuvers. Kuge et al.\cite{kuge2000driver} focused on modeling emergency and normal lane changes, employing customized HMMs for these specific scenarios. To capture the temporal information inherent in driving behaviors, Recurrent Neural Network (RNN) based methods\cite{xu2017end, xu2019temporal} were proposed. Xu et al.\cite{xu2017end} utilized an FCN-LSTM architecture, where convolution neural networks extract image features and LSTM models the temporal relationships. Additionally, Xu et al.\cite{xu2019temporal} introduced a Temporal Recurrent Network that uses LSTM to perform both online and future action prediction, leveraging historical evidence and predicted future information for better current action recognition. Guo et al.\cite{guo2022uncertainty}introduces an uncertainty-based spatiotemporal attention mechanism for online action detection. By incorporating a probabilistic model into the baseline framework, it quantifies prediction uncertainty to generate spatiotemporal attention. With the advent of Transformer models demonstrating powerful modeling capabilities in computer vision, researchers have designed various Transformer architectures for end-to-end online action detection\cite{noguchi2023ego, cao2022circular}. Colar\cite{yang2022colar} utilizes an exemplar-query mechanism for online action detection. This mechanism initially compares the similarity between the current frames and exemplar frames, and subsequently aggregates the exemplar features weighted by their similarity. GateHUB\cite{chen2022gatehub} enhances the accuracy of online action detection by employing Gated History Units (GHU) and Future-augmented History (FaH). E2E-LOAD\cite{noguchi2023ego} employs a "spatial-temporal" paradigm to model long and short-term dependencies, achieving efficient temporal representation. CWC-Trans\cite{cao2022circular} introduces a cascaded Transformer architecture based on a cyclic window and a recurrent historical queue, implementing multi-stage attention and cascaded refinement for each window. Unlike these existing methods, we set our sights on the solving ambiguity and similarity hidden in different driving scenes using proposed self-discovery framework.

\begin{figure*}[htbp]
  \centering
  \includegraphics[width=\textwidth]{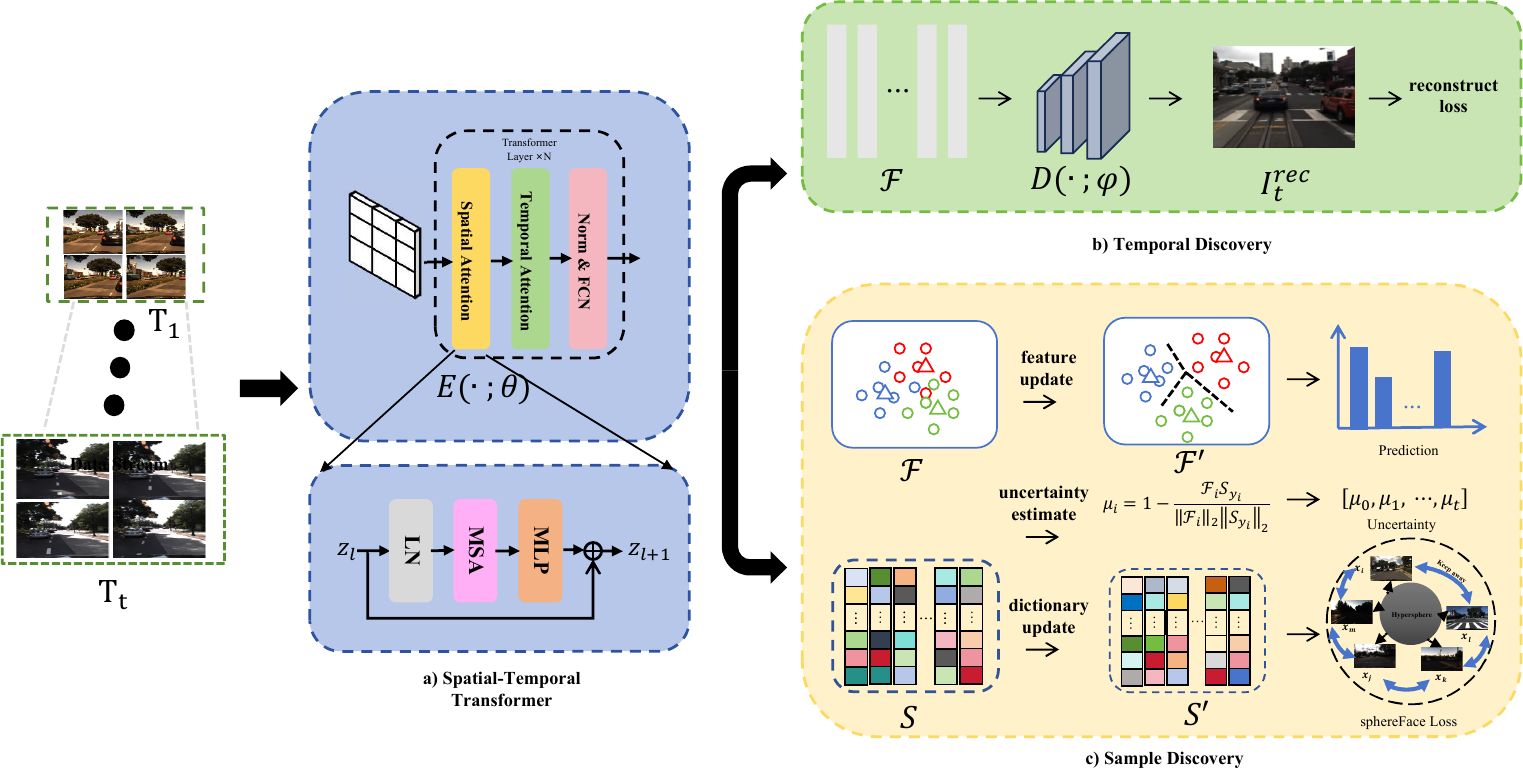}
  \caption{The proposed framework consists of three parts: (1) spatial-temporal Transformer for feature extraction, (2) temporal discovery module for feature representation \( \mathcal{F} \) reconstruction using decoder \( D(\cdot; \varphi) \), and (3) sample discovery module to reduce scene confusion with dynamically updated meta-variables and input feature weights.}
  \label{fig:framework}
\end{figure*}

\subsection{Self-Supervised Learning}
Self-supervised learning leverages the inherent structure and properties of data to generate pseudo-labels, enabling model training without manual annotations and facilitating the learning of robust representations. Given the strong capability of self-supervised tasks in extracting high-level semantics, many researchers have adapted this approach to the autonomous driving domain\cite{luo2021self, sautier2022image,bhattacharyya2023ssl,chen2021multisiam}. For instance, Luo et al.\cite{luo2021self} assumes that the structure of objects remains consistent between consecutive point cloud scans and learns motion by comparing the structural similarity between transformed and real point clouds. Sautier et al.\cite{sautier2022image} employs contrastive learning by matching 3D point features with corresponding 2D image features. Bhattacharyya et al.\cite{bhattacharyya2023ssl} learns local structural information by randomly masking parts of lane map features and requiring the model to reconstruct these features. Chen et al.\cite{chen2021multisiam} performs clustering within a single image rather than across the entire dataset to further explore intra-image similarities and enhance the translational invariance of the learned representations.

\subsection{Dictionary Learning}

Dictionary learning is a method that represents data samples sparsely by learning a set of atoms (dictionary) that effectively capture the data structure and features. In recent years, dictionary learning has shown outstanding performance in various fields, including image denoising\cite{zheng2021deep}, image classification\cite{shao2020label,tang2020dictionary}, graph representation learning\cite{zeng2023generative} and so on. Zheng et al.\cite{zheng2021deep} models the representation, learns the priors of the representation coefficients and dictionary, can adaptively adjust the dictionary based on the input image content. Zeng et al.\cite{zeng2023generative} tackles the graph dictionary learning problem from a generative angle, using graph generation functions based on Radial Basis Function (RBF) kernels and Fused Gromov-Wasserstein (FGW) distance to create nonlinear embedding spaces that closely approximate the original graph space. Some researchers\cite{shao2020label,tang2020dictionary} apply dictionary learning to image classification tasks, sparsifying information from different categories to derive distinguishable feature representations. Inspired by the robust modeling capability of dictionary learning for samples and its strong representation ability in the category space, we establish dictionary sets for different categories and update the dictionaries and features in real-time during the training phase to accurately represent different driving behaviors from the feature basis space.

\section{Method}

This paper aims to propose a concise framework based on self discovery learning for driving behaviors recognition. The proposed SDL does not require any external supervision or additional manual labels. By adding SDL into the existing architectures, the network can learn to strengthen its representations and fully distinguish the confused information of input sequences.

Fig.\ref{fig:framework} illustrates the framework of our proposed SDL, which consists of three parts: (1) spatial-temporal transformer, (2) temporal discovery module, (3) sample discovery module. Inspired by the success of the transformer in auto-driving, we adopt the spatial-temporal transformer as a feature extractor \( E(\cdot; \theta) \) to obtain a strong representation \( \mathcal{F} \) for input video sequence. Then, the extracted feature representation \( \mathcal{F} \) is reconstructed into the original image \( I_{rec} \) using the decoder \( D(\cdot; \varphi) \). It should be noted that the decoder is utilized solely during the training phase and does not add any extra parameters to the model. To reduce the confusion among different scenes, inspired by dictionary learning, meta-variables are used to represent the feature vectors of different classes, both the features and the dictionary are dynamically updated in the training process. Furthermore, we adjust the weights of the input sequence accordingly by estimating uncertainty through the distance between meta-variables and input features. Below, we present the details of the three parts in SDL.

\subsection{Spatial-Temporal Transformer}

Traditional Vision Transformers (ViT) extract \(N\) non-overlapping image patches \(x_i \in \mathbb{R}^{h \times w \times c}\) from an image, which are then linearly projected into one-dimensional vectors \(z_i \in \mathbb{R}^d\) and fed into the Transformer encoder. In contrast, temporal transformers use Tubelet embedding. For an input video \(T \in \mathbb{R}^{t \times h \times w \times c}\), embeddings \(z \in \mathbb{R}^{n_t \times n_h \times n_w \times d}\) are obtained via 3D convolution, extracting features from temporal, height, and width dimensions. 

In embedding phase, the process starts with dividing the input video \(V\) into non-overlapping tubelets of size \(P \times P \times P_t\), where \(P\) is the spatial dimension and \(P_t\) is the temporal dimension. Each tubelet is then flattened and linearly projected into a vector \(z_i \in \mathbb{R}^d\). The positional embeddings are added to these vectors to retain spatial and temporal positional information. The resulting embeddings are reshaped into \(\mathbb{R}^{N \times d}\) for input to the transformer encoder, where \(N = \frac{T \cdot H \cdot W}{P_t \cdot P \cdot P}\).

The video transformer network introduces a spatiotemporal attention mechanism, first computing spatial self-attention among tokens at the same temporal index and then temporal self-attention among tokens at the same spatial index. The feature vectors are reshaped to  \(z \in \mathbb{R}^{n_t \times n_h \times n_w \times d}\) into \(z_s \in \mathbb{R}^{n_t \times (n_h \cdot n_w) \times d}\) for spatial attention, and then \(z_t \in \mathbb{R}^{(n_t \cdot n_h) \times n_w \times d}\) for temporal attention. The attention mechanism is defined as follows:
\begin{align}
y_s^l &= \text{MSA}(\text{LN}(z_s^l)) + z_s^l \\
y_t^l &= \text{MSA}(\text{LN}(y_s^l)) + y_s^l \\
z^{l+1} &= \text{MLP}(\text{LN}(y_t^l)) + y_t^l
\end{align}
where MSA stands for multi-head self-attention, MLP for multi-layer perceptron, and LN for layer normalization.

In the spatial self-attention, the attention weights are computed among tokens at the same temporal index:
\begin{align}
\text{Attention}(Q, K, V) = \text{softmax}\left(\frac{QK^T}{\sqrt{d_k}}\right)V
\end{align}
where \(Q = W_Q z_s\), \(K = W_K z_s\), and \(V = W_V z_s\) are the query, key, and value matrices, respectively, and \(d_k\) is the dimension of the key vectors.

Similarly, for the temporal self-attention, the attention weights are computed among tokens at the same spatial index. The features integrated by temporal attention are finally classified using MLP. This framework is suitable for both frame classification and frame prediction tasks. For frame classification, the features \(z_t\) are reshaped into \(\mathbb{R}^{T \times (n_h \cdot n_w \cdot d)}\) and predicted using \(T\) MLP heads. For frame prediction, the classification head \(z_{\text{cls}}\) is directly used for MLP classification.

To summarize, the spatial-temporal transformer effectively captures both spatial and temporal dependencies in video data, providing a strong representation for driving behavior recognition. This enhances the model's ability to distinguish between different driving scenarios and improves overall performance.

\subsection{Temporal Discovery}

As previously mentioned, distinguishing between confusing behaviors requires capturing additional information ,such as background details and sequence differences, to enhance the differentiation. Intuitively, temporal changes are more effective than single-frame differences for distinguishing input samples. To enhance the model's ability to capture temporal patterns and obtain feature representations with better temporal generalization, we introduce a self-supervised learning framework to discover temporal information. As illustrated in Fig. ??????\ref{fig:framework} b), the feature representation \( \mathcal{F} \) captured by the spatiotemporal transformer is fed into a decoder \( D(\cdot; \phi) \). The decoder's role is to transform the high-dimensional feature representation back into the original image space. Then, we define a reconstruction loss that constrains the relationship between the reconstructed image and the original input to encourage the network to recognize and emphasize temporal differences, formally:
\begin{align}
    \mathcal{L}_{\text{rec}} = \frac{1}{N} \sum_{i=1}^{N} \left\| I_t^{(i)} - I^{\text{rec}(i)} \right\|^2
\end{align}
The objective of this constraint is to directly minimize the discrepancy between the reconstructed image and the original image. It is worth noting that our temporal discovery module focuses on reconstructing the last frame of the image sequence instead of predicting unseen future frames. This approach is chosen because the input sequence lacks future frame information, and the reconstruction task is sufficient to capture the motion information effectively\cite{wang2022repre}. By learning through the self-supervised task, the network becomes capable of perceiving sequence differences and providing additional information, thereby achieving more precise classification. Moreover, this module requires no extra manual annotations and does not introduce additional network parameters during the inference phase, which is advantageous for learning with small-scale datasets and maintaining real-time network performance.

\begin{algorithm}[H]
\caption{Sample Discovery}
\label{alg:online_dict_learning}
\begin{algorithmic}[1]
\REQUIRE Initial dictionary $\mathcal{S}$, Input feature $\{\mathcal{F}_i\}_{i=1}^N$, label $\{y_i\}_{i=1}^N$, feature update rate $\alpha$, dictionary update rate $\beta$
\ENSURE Updated dictionary $\mathcal{S}'$, Updated feature $\{\mathcal{F}_i'\}_{i=1}^N$
\STATE Initialize dictionary $\mathcal{S}$
\FOR{each feature $\mathcal{F}_i$ in sequence}
    \STATE \textbf{Feature Update:}
    \STATE Let $\mathcal{S}_{y_i}$ be the dictionary atom corresponding to label $y_i$
    \STATE Update feature $\mathcal{F}_i$ as in Equ.\eqref{equ: feature_update}
    
    \STATE \textbf{Uncertainty Estimate:}
    \STATE Compute the cosine distance between $\mathcal{F}_i'$ and $\mathcal{S}_{y_i}$ as in Equ.\eqref{equ:cosine_distance}
    \STATE Compute the weight $w_i$ as in Equ.\eqref{equ:w_i}
    
    \STATE \textbf{Dictionary Update:}
    \STATE Let $\mathcal{S}_{y_i}$ be the dictionary atom corresponding to label $y_i$
    \STATE Update dictionary $\mathcal{S}_{y_i}$ as in Equ.\eqref{equ:dictionary_update}
\ENDFOR
\STATE \textbf{return} Updated dictionary $\mathcal{S}'$ , updated features $\{\mathcal{F}_i'\}_{i=1}^N$ and sample weight $\{w_i\}_{i=1}^N$
\end{algorithmic}
\end{algorithm}

\subsection{Sample Discovery}

To capture subtle differences across various scenes and help the network correctly distinguish the potential semantic differences between similar categories, we drew inspiration from dictionary learning and designed an online dictionary learning framework for sample discovery. This framework increases the distance between meta-vectors in spherical space and dynamically updates feature representations and the dictionary set in real time, thereby enhancing inter-class differences across different scenes. Furthermore, this module estimates the uncertainty of the input features to resample the training samples. The detailed algorithm is presented in Alg.\ref{alg:online_dict_learning}.

\textbf{Feature Update: }
The basis vectors $\mathcal{S}_i$ in the dictionary set represent the common features of different classes on the classification sphere, encapsulating their unique semantic information. Conversely, the feature vectors $\mathcal{F}$ derived from the spatio-temporal transformer are not sensitive to subtle differences between categories. Therefore, we perform a weighted sum of the original features $\mathcal{F}$ and the basis vectors $\mathcal{S}_{y_i}$ corresponding to their categories, formally:
\begin{align}
    \mathcal{F}_i' \leftarrow (1 - \alpha) \mathcal{F}_i + \alpha \mathcal{S}_{y_i}
\label{equ: feature_update}
\end{align}
where $\mathcal{S}_{y_i}$ denotes the meta-vector for the corresponding category ${y_i}$, and $\alpha$ is the weight which is set 0.9 in our experiments. It is important to note that the update of features depends on stable dictionary feature representations. Hence, in practice, we schedule the feature update to occur after epoch 10.

\textbf{Uncertainty Estimate: }
As mentioned above, scenario confusion and annotation ambiguity are prevalent in training samples. For instance, both left lane change and left lane branch involve left-turn behavior, crosswalk passing samples include scenarios of intersection passing. Intuitively, forcing the model to distinguish these confusing samples is risky and could cause the model to fall into local optima specific to the training scenarios. Conversely, leveraging the strong descriptive power of dictionary vectors for class space, we propose an uncertainty estimation algorithm. Specifically, we use dictionary vectors as a reference and compute the cosine distance between the feature vector and the dictionary vector of the corresponding category to measure the feature's uncertainty, formally:
\begin{align}
     \mu_i = 1 - \frac{\mathcal{F}_i \cdot \mathcal{S}_{y_i}}{\|\mathcal{F}_i\|_2 \|\mathcal{S}_{y_i}\|_2}
\label{equ:cosine_distance}
\end{align}
The uncertainty index $\mu_i$ will be normalized and used as the classification weight for the current frame. This means that features closer to the base vector contribute more to the network parameter updates, while those further away contribute less. The specific formula is as follows:
\begin{align}
    w_i = \text{clamp}(1 - \mu_i, 0.5, 1)
    \label{equ:w_i}
\end{align}
In line with the feature update, the uncertainty estimation is conducted only after achieving a stable dictionary feature representation.

\textbf{Dictionary Update: }In our framework, we utilize the learned parameter dictionary $\mathcal{S} \in \mathbb{R}^{C \times D}$ to represent each driving behavior base vector. The dictionary set $\mathcal{S}$ is randomly initialized and dynamically updated during the network's learning process, capturing and modeling the distributions of different categories in real-time, formally:
\begin{align}
    \mathcal{S}_{y_i}' \leftarrow (1 - \beta) \mathcal{S}_{y_i} + \beta \mathcal{F}_i
\label{equ:dictionary_update}
\end{align}
where $\beta$ is the hyper-parameter, gradually decreases as the training progresses, that controls the trade-off between the original dictionary  vector $\mathcal{S}_{y_i}$ and the learned feature $\mathcal{F}_i$. Additionally, we incorporate the SphereFace loss\cite{liu2017sphereface} to regulate the distribution space of the dictionary set. Formally:
\begin{align}
L_{dict} = -\frac{1}{C} \sum_{i=1}^C \log \frac{e^{s (\cos(m \theta_{y_i}) - \delta)}}{e^{s (\cos(m \theta_{y_i}) - \delta)} + \sum_{j \neq y_i} e^{s \cos \theta_j}}
\end{align}

where $s$ and $m$ denote for scale and margin, respectively. The goal of this constraint is to improve the inter-class distances of the dictionary atoms, which in turn helps to make the features more discriminative.

\subsection{Objective function}
Apart from the mentioned losses, we incorporate the conventional classification loss. The joint objective function can be formulated as:
\begin{align}
    \mathcal{L} =  \sum_{i=1}^Nw_i\times\mathcal{L}_{\text{ce}_i} + \lambda_1 \times \mathcal{L}_{\text{rec}} +  \lambda_2 \times \mathcal{L}_{\text{dict}}
\end{align}
where the $\mathcal{L}_{\text{ce}_i}$ is the the standard cross-entropy loss which use the ground-truth as a supervised signal to guide the classification learning process, $w$ is the loss weight computed from Equ.\ref{equ:w_i}. The symbol $\lambda$ is a hyperparameter that controls the importance of the extra loss.

\section{Experiment}
\subsection{Implementation Details}
\textbf{Dataset:} The Honda Research Institute Driving Dataset (HDD) was utilized to assess the proposed method. This dataset comprises 104 hours of first-person driving videos captured in the San Francisco Bay Area, featuring detailed frame-level annotations for various vehicle actions. The video resolution is 1280 × 720, with a frame rate of 30 frames per second. Consistent with previous studies, the dataset includes labels for 11 distinct Goal-oriented actions, such as left turn, right lane change, merge, and others. Additionally, there are 6 Cause labels, which encompass five types of stopping actions based on their causes (e.g., congestion, sign, red light) and one deviating action. Following the approach in prior research\cite{cao2023e2e}, the 137 sessions in the HDD were divided into 100 training sessions and 37 testing sessions. 

\textbf{Training:} The input images are resized to a fixed size of 256 × 256 and randomly cropped into 224 × 224. spatial and temporal augmentation, such as randomly cropped, color jitter, temporal jitter, is exploited before feature extraction. During the training process, the category label of the input frame is the only manual annotation used for training. We implemented all the expriments using Pytorch\cite{paszke2017automatic}. The loss weight $\lambda$ is set 0.1 and 0.5 for the reconstruct loss and dictionary loss respectively. We train the whole model for 100 epochs with a cosine decay learning rate scheduler using the AdamW\cite{loshchilov2017decoupled} optimizer. To showcase the advantages of our proposed SDL method, we employ E2E-load as a feature extractor and integrate SDL into the training process.

\textbf{Inference:} During the inference phase, we merely input the original video sequence into the trained model and SDL framework is unnecessary. It means the proposed method does not introduce computational overhead at inference time.

\textbf{Evaluation Metric:} We employ the mean Average Precision (mAP) metric to assess the performance of the proposed method. AP, the area under the precision-recall curve, is a standard evaluation method used in previous research.

\begin{table*}[ht]
\centering
\scriptsize
\caption{Comparison of classification performance between the proposed and existing methods for labels of 11 Goal-oriented actions. Following the original paper reports, some driving behavior APs are not shown. * indicates reproduce result.}
\begin{tabular}{p{1.8cm}p{1cm}p{0.8cm}p{0.8cm}p{0.8cm}p{0.8cm}p{0.8cm}p{0.8cm}p{0.8cm}p{0.8cm}p{0.8cm}p{0.8cm}p{0.8cm}}
\toprule
Methods & intersection \newline passing & L \newline turn & R \newline turn & L lane \newline change & R lane \newline change & L lane \newline branch & R lane \newline branch & crosswalk \newline passing & railroad \newline passing & merge & u-turn & Overall \newline mAP \\ \midrule
CNN\cite{ramanishka2018toward} & 53.4 & 47.3 & 39.4 & 23.8 & 17.9 & 25.2 & 2.9 & 4.8 & 1.6 & 4.3 & 7.2 & 20.7 \\ 
CNN-LSTM\cite{ramanishka2018toward} & 65.7 & 57.7 & 54.4 & 27.8 & 26.1 & 25.7 & 1.7 & 16.0 & 2.5 & 4.8 & 13.6 & 26.9 \\ 
ED\cite{xu2019temporal} & 63.1 & 54.2 & 55.1 & 28.3 & 35.9 & 27.6 & 8.5 & 7.1 & 0.3 & 4.2 & 14.6 & 27.2 \\ 
TRN\cite{xu2019temporal} & 63.5 & 57.0 & 57.3 & 28.4 & 37.8 & 31.8 & 10.5 & 11.0 & 0.5 & 6.3 & 16.7 & 33.7 \\ 
DEPSEG-LSTM\cite{narayanan2018semi} & 70.9 & 63.4 & 63.6 & 48.0 & 40.9 & 39.7 & 4.4 & 16.1 & 0.5 & 6.3 & 16.7 & 33.7 \\ 
C3D\cite{tran2015learning} & 72.8 & 64.8 & 71.7 & 53.4 & 44.7 & 52.2 & 3.1 & 14.6 &  2.9 & 10.6 & 15.8 & 37.0 \\ 
OadTR\cite{wang2021oadtr} & 44.4 & 73.0 & \textbf{75.9} & 33.3 & 23.5 & 19.8 & 0.02 & 0.007 & 0.001 & 0.03 & 0.002 & \text{25.1} \\
Colar\cite{yang2022colar} & - & - & - & - & - & - & - & - & - &- & - & 30.6 \\
GateHUB\cite{chen2022gatehub} & - & - & - & - & - & - & - & - & - & - & - & 32.1 \\
Uncertain-OAD\cite{guo2022uncertainty} &  - & - & - & - & - & - & - & - &- & - & - & 30.1 \\
E2E-Load\cite{cao2023e2e} & 84.8 & 72.8 & 72.7 & 58.2 & \textbf{56.9} & 59.6 & 14.6 & 39.0  & \textbf{9.6} & 30.3 & \textbf{25.5} & \textbf{47.6} \\
SDL(our)  & \textbf{89.7} & \textbf{74.8} & 73.4 & \textbf{65.6} & 55.9 & \textbf{65.6} & \textbf{17.0} & \textbf{39.4} &2.4 & \textbf{32.0} & 24.3 & \textbf{49.1} \\
\bottomrule
\end{tabular}
\label{table:comparison}
\end{table*}

\subsection{Comparison with State-of-the-art methods}

We compare our proposed SDL with state-of-the-art action recognition methods on the HDD dataset. As illustrated in Tab.ref{table:comparison}, our proposed SDL achieve the considerable performance, outperforms the other methods by 1\% in mAP. It has reached the best or second-best performance across various driving scenarios, particularly in low-sample or difficult-to-distinguish situations such as merging and crosswalk passing, showing significant improvements in AP. Conventional CNN-LSTM architectures\cite{ramanishka2018toward, xu2019temporal, narayanan2018semi} tend to process spatial and temporal features sequentially, which inherently limits their ability to effectively capture temporal information. Subsequent approaches, such as the C3D method\cite{tran2015learning}, utilize 3D convolutions to address this limitation by simultaneously handling spatial-temporal information. Transformer architectures\cite{yang2022colar, cao2023e2e}, on the other hand, naturally complement long sequence processing, enabling them to capture subtle differences in temporal context. Our proposed SDL method leverages its intrinsic capabilities, employing two modules—temporal discovery and sample discovery—to capture more robust spatial-temporal features. This approach allows for the filtering of confusing samples, thereby ensuring accurate judgments across various driving behavior scenarios.

\subsection{Ablation Study}

\begin{table}[]
    \centering
    \caption{SDL Ablation Study on HDD. TD and SD denote Temporal Discovery and Sample Discovery module respectively. Bold denotes the best.}
    \begin{tabular}{cccc}
        \toprule
         baseline & TD & SD & mAP   \\
         \midrule
          \checkmark & & &  47.6 \\
          \checkmark & \checkmark & &  48.4 \\
          \checkmark & & \checkmark &  48.8 \\
          \checkmark & \checkmark & \checkmark &  \textbf{49.1} \\
          \bottomrule
    \end{tabular}
    \label{tab:ablation}
\end{table}

\begin{figure*}[htbp]
  \centering
  \includegraphics[width=\textwidth]{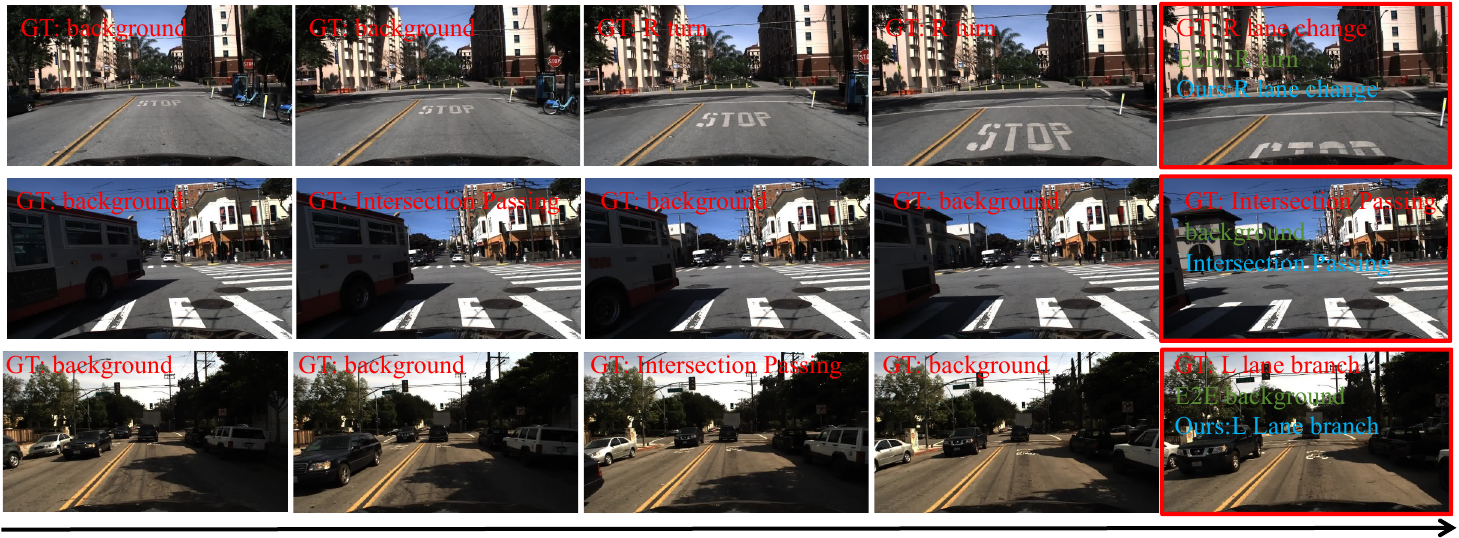}
  \caption{Online detection examples on the HDD dataset comparing our proposed SDL method with the baseline E2E-load. Red boxes indicate retrieval frames, red text denotes ground truth, green text represents E2E predictions, and blue text shows SDL results. Our method effectively distinguishes similar behaviors and improves accuracy in scenarios with short-duration actions, demonstrating superior spatial-temporal perception and hidden cue capture.}
  \label{fig:vis}
\end{figure*}

\textbf{Contributions of different modules in SDL: }To analyze the contribution of each module, we conduct comprehensive ablation experiments on HDD dataset. As illustrated in Tab.\ref{tab:ablation}, each proposed module contributes improvement for the recognition performance. With the contribution of temporal discovery module, the recognition achieves 0.3\% mAP. The improvement indicates that the model could understand and perceive hidden temporal cue with reconstruct task. With the proposed sample discovery task, the performance obtains 0.7\%. It is because the confused sample has been filtered and model build more differentiated hyperspace by dictionary learning. What's more, combined with temporal discovery and sample discovery module, can obtains 1\% improvements which demonstrates different modules can synergistically enhance each other.In summary, each proposed module is effective and beneficial for recognition performance.
\begin{figure}[htbp]
  \centering
  \includegraphics[width=\columnwidth]{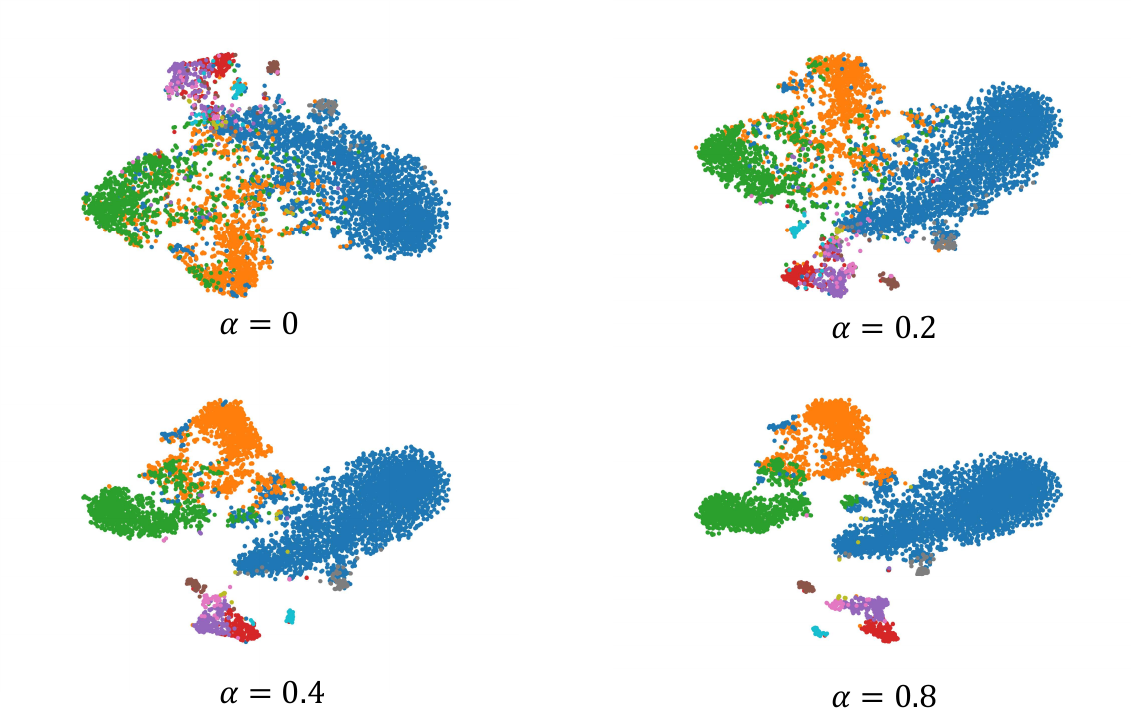}
  \caption{Visualization of frame-level features using t-SNE with varying $\alpha$ values in Equation \ref{equ: feature_update}. As $\alpha$ increases, inter-class distances expand, enhancing feature discriminative power.}
  \label{fig:cluster}
\end{figure}

\begin{figure}[htbp]
  \centering
  \includegraphics[width=\columnwidth]{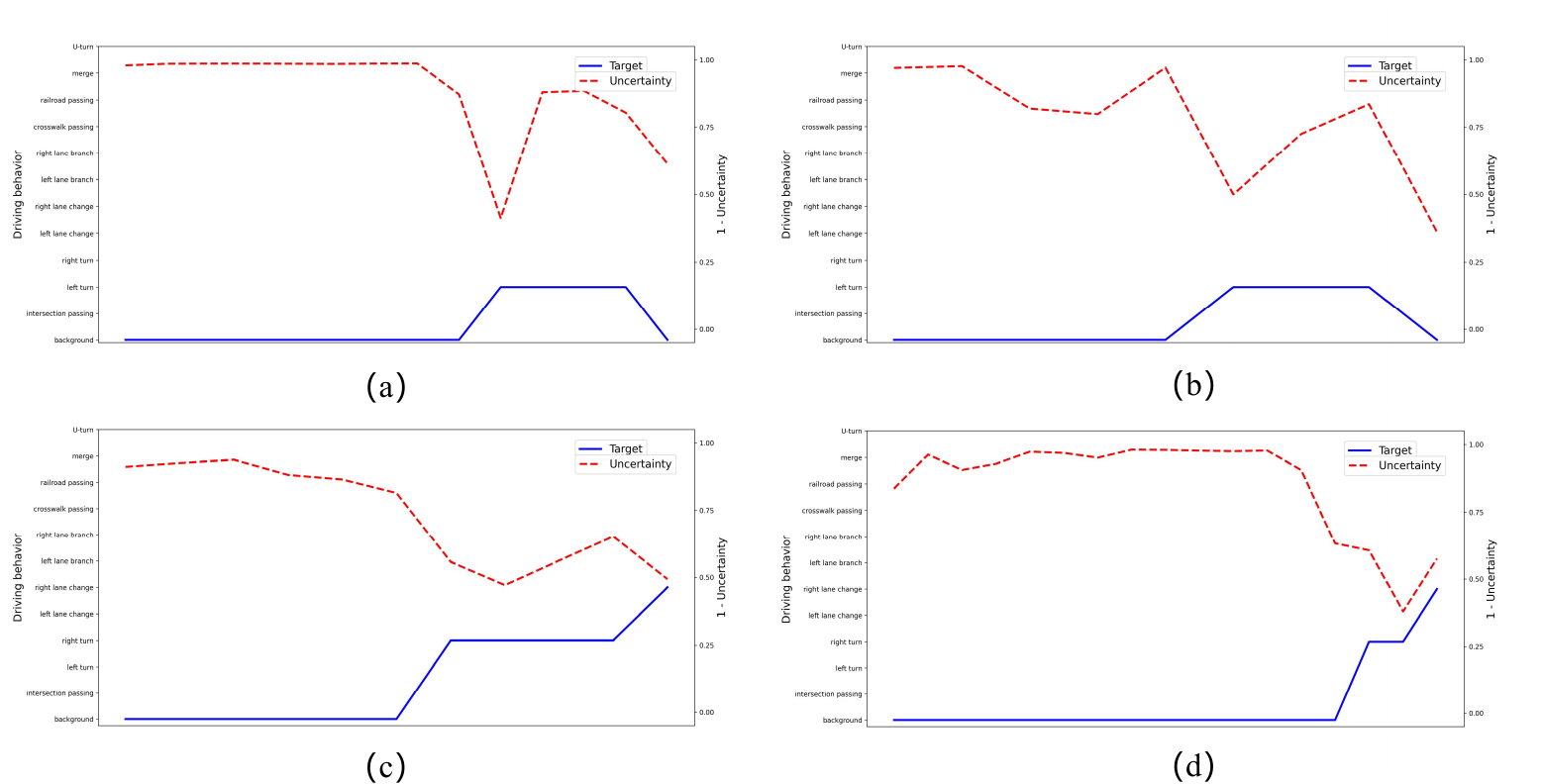}
  \caption{Visualization of driving behavior changes and corresponding uncertainty curves over time. Consistent behavior shows high, stable uncertainty, while behavior changes cause sharp drops in uncertainty, indicating boundary frames between behaviors.}
  \label{fig:uncertainty}
\end{figure}

\textbf{Analysis of the choice of the self-supervised task:}

\textbf{Analysis of the dictionary set: } We examined the function of the dictionary set learned within the sample discovery module. We extracted frame-level features from all test sets (excluding the "background" class) and visualized these features using the t-SNE method. Figure \ref{fig:cluster} illustrates the results under different $\alpha$ values as defined in Equation \ref{equ: feature_update}. As shown in Figure \ref{fig:cluster} a), it is evident that the distribution in the original category space is compact, making it easy to confuse samples from different categories. As $\alpha$ increases, meaning more reference information is derived from the category meta-vectors of the dictionary set, the inter-class distances expand, enhancing the discriminative power of the features. This demonstrates that: 1) the dictionary vectors learned through sample discovery effectively represent the distribution of different categories; 2) the dynamic update learning strategy for dictionary vectors enables the network to improve its perception of different driving behavior categories.

\textbf{Analysis of the uncertainty estimate: } To illustrate the effectiveness of the uncertainty derived from sample discovery, we sampled different video clips and visualized the changes in driving behavior over time along with the corresponding uncertainty curves of the frames. As shown in Fig. \ref{fig:uncertainty} a) and b), it is evident that when the driving behavior remains consistent, the uncertainty stays high and relatively stable. This suggests that the model's understanding of the sample category is clear and unambiguous. However, when the driving behavior changes, the uncertainty drops sharply. This occurs because the current frame lies at the boundary between two different driving behaviors, meaning such samples may exhibit characteristics of both scenarios. Forcing the model to make a hard classification into a single category in such cases can be detrimental to its learning process. Our proposed SDL method dynamically estimates the uncertainty of such samples and applies a lower weight for learning, thereby reducing the negative impact of confusing scenes. Notably, we also observed that frequent action changes lead to sustained fluctuations in uncertainty, maintaining a lower level, as shown in Fig. \ref{fig:uncertainty} c) and d). This indicates a decrease in the model's discrimination ability in such scenarios, likely due to the confusion of similar scenes, which is an area worth exploring further in the future.

\subsection{Visualization}
We further give the visualization of our proposed SDL to demonstrate the superiority compared baseline E2E-load. As shown in Fig.\ref{fig:vis}, We present online detection examples on the HDD dataset. Red boxes indicate retrieval frames, red text denotes ground truth, green text represents E2E prediction results, and blue text shows the results of our proposed SDL. For simplicity, we only display the most recent 5 frames.

In the first row, we observe that the input sequence includes very similar driving behaviors, such as "R turn" and "R lane change." These challenging scenarios require the model to rely heavily on contextual information and clearly differentiate between the two driving behaviors to make accurate judgments. The E2E-load model lacks this capability, whereas our method successfully addresses this challenge. Furthermore, in real-world scenarios, vehicle behaviors are not always continuous, and the duration of a driving behavior can sometimes be very short. Models often tend to predict the same action as the previous frame, leading to inaccuracies at the onset of a new action. However, by proactively mining temporal and sample information, our method has made significant improvements in this regard.The visualization results in Fig.\ref{fig:vis} verify the effectiveness of our proposed method that can perceive the strong spatial-temporal variation and capture hidden cue through an elegant self discovery framework.

\section{Conclusion}
In this paper, we propose a Self Discovery Learning framework to enhance the recognition of driving behaviors by addressing the challenges of sample scarcity and confusion. The fundamental concept of our proposed SDL is to capture subtle variations in driving behavior through in-depth exploration of intrinsic patterns and to handle different confusing samples distinctively. Our method utilize the spatial-temporal transformer as feature exactor to obtain robust feature representation. Inspired by the strong capability of self-supervised tasks in modeling deep semantic features, we construct a sub-pixel level reconstruction learning framework for input sequences.Furthermore, we establish a set of dictionary vectors to represent different categories in high-dimensional space. By leveraging the metric relationship between the dictionary vectors and the extracted features, we dynamically update the features and the dictionary in real-time, aiming to make the feature space of different categories more distinct at the classification boundary. Additionally, utilizing the benchmark role of the dictionary vectors, we estimate the uncertainty of the input frame features as a measure of sample confusion. The quantitative and qualitative experimental results demonstrate the effectiveness and superiority of our proposed SDL. Future work will explore strategies to mitigate the impact of the long-tail distribution on model performance and develop methods to better differentiate subtle variations in driving behaviors under few-shot learning conditions.

\bibliographystyle{IEEEtran}
\bibliography{IEEEabrv, reference}

\begin{thebibliography}{10}
\providecommand{\url}[1]{#1}
\csname url@rmstyle\endcsname
\providecommand{\newblock}{\relax}
\providecommand{\bibinfo}[2]{#2}
\providecommand\BIBentrySTDinterwordspacing{\spaceskip=0pt\relax}
\providecommand\BIBentryALTinterwordstretchfactor{4}
\providecommand\BIBentryALTinterwordspacing{\spaceskip=\fontdimen2\font plus
\BIBentryALTinterwordstretchfactor\fontdimen3\font minus \fontdimen4\font\relax}
\providecommand\BIBforeignlanguage[2]{{%
\expandafter\ifx\csname l@#1\endcsname\relax
\typeout{** WARNING: IEEEtran.bst: No hyphenation pattern has been}%
\typeout{** loaded for the language `#1'. Using the pattern for}%
\typeout{** the default language instead.}%
\else
\language=\csname l@#1\endcsname
\fi
#2}}

\bibitem{kaplan2015driver}
S.~Kaplan, M.~A. Guvensan, A.~G. Yavuz, and Y.~Karalurt, ``Driver behavior analysis for safe driving: A survey,'' \emph{IEEE Transactions on Intelligent Transportation Systems}, vol.~16, no.~6, pp. 3017--3032, 2015.

\bibitem{ramanishka2018toward}
V.~Ramanishka, Y.-T. Chen, T.~Misu, and K.~Saenko, ``Toward driving scene understanding: A dataset for learning driver behavior and causal reasoning,'' in \emph{Proceedings of the IEEE Conference on Computer Vision and Pattern Recognition}, 2018, pp. 7699--7707.

\bibitem{noguchi2023ego}
C.~Noguchi and T.~Tanizawa, ``Ego-vehicle action recognition based on semi-supervised contrastive learning,'' in \emph{Proceedings of the IEEE/CVF Winter Conference on Applications of Computer Vision}, 2023, pp. 5988--5998.

\bibitem{cao2023e2e}
S.~Cao, W.~Luo, B.~Wang, W.~Zhang, and L.~Ma, ``E2e-load: end-to-end long-form online action detection,'' in \emph{Proceedings of the IEEE/CVF International Conference on Computer Vision}, 2023, pp. 10\,422--10\,432.

\bibitem{dosovitskiy2020image}
A.~Dosovitskiy, ``An image is worth 16x16 words: Transformers for image recognition at scale,'' \emph{arXiv preprint arXiv:2010.11929}, 2020.

\bibitem{jing2020self}
L.~Jing and Y.~Tian, ``Self-supervised visual feature learning with deep neural networks: A survey,'' \emph{IEEE transactions on pattern analysis and machine intelligence}, vol.~43, no.~11, pp. 4037--4058, 2020.

\bibitem{kuge2000driver}
N.~Kuge, T.~Yamamura, O.~Shimoyama, and A.~Liu, ``A driver behavior recognition method based on a driver model framework,'' \emph{SAE transactions}, pp. 469--476, 2000.

\bibitem{jain2015car}
A.~Jain, H.~S. Koppula, B.~Raghavan, S.~Soh, and A.~Saxena, ``Car that knows before you do: Anticipating maneuvers via learning temporal driving models,'' in \emph{Proceedings of the IEEE International Conference on Computer Vision}, 2015, pp. 3182--3190.

\bibitem{oliver2000graphical}
N.~Oliver and A.~P. Pentland, ``Graphical models for driver behavior recognition in a smartcar,'' in \emph{Proceedings of the IEEE Intelligent Vehicles Symposium 2000 (Cat. No. 00TH8511)}.\hskip 1em plus 0.5em minus 0.4em\relax IEEE, 2000, pp. 7--12.

\bibitem{xu2017end}
H.~Xu, Y.~Gao, F.~Yu, and T.~Darrell, ``End-to-end learning of driving models from large-scale video datasets,'' in \emph{Proceedings of the IEEE conference on computer vision and pattern recognition}, 2017, pp. 2174--2182.

\bibitem{xu2019temporal}
M.~Xu, M.~Gao, Y.-T. Chen, L.~S. Davis, and D.~J. Crandall, ``Temporal recurrent networks for online action detection,'' in \emph{Proceedings of the IEEE/CVF international conference on computer vision}, 2019, pp. 5532--5541.

\bibitem{guo2022uncertainty}
H.~Guo, Z.~Ren, Y.~Wu, G.~Hua, and Q.~Ji, ``Uncertainty-based spatial-temporal attention for online action detection,'' in \emph{European Conference on Computer Vision}.\hskip 1em plus 0.5em minus 0.4em\relax Springer, 2022, pp. 69--86.

\bibitem{cao2022circular}
S.~Cao, W.~Luo, B.~Wang, W.~Zhang, and L.~Ma, ``A circular window-based cascade transformer for online action detection,'' \emph{arXiv preprint arXiv:2208.14209}, 2022.

\bibitem{yang2022colar}
L.~Yang, J.~Han, and D.~Zhang, ``Colar: Effective and efficient online action detection by consulting exemplars,'' in \emph{Proceedings of the IEEE/CVF conference on computer vision and pattern recognition}, 2022, pp. 3160--3169.

\bibitem{chen2022gatehub}
J.~Chen, G.~Mittal, Y.~Yu, Y.~Kong, and M.~Chen, ``Gatehub: Gated history unit with background suppression for online action detection,'' in \emph{Proceedings of the IEEE/CVF Conference on Computer Vision and Pattern Recognition}, 2022, pp. 19\,925--19\,934.

\bibitem{luo2021self}
C.~Luo, X.~Yang, and A.~Yuille, ``Self-supervised pillar motion learning for autonomous driving,'' in \emph{Proceedings of the IEEE/CVF Conference on Computer Vision and Pattern Recognition}, 2021, pp. 3183--3192.

\bibitem{sautier2022image}
C.~Sautier, G.~Puy, S.~Gidaris, A.~Boulch, A.~Bursuc, and R.~Marlet, ``Image-to-lidar self-supervised distillation for autonomous driving data,'' in \emph{Proceedings of the IEEE/CVF Conference on Computer Vision and Pattern Recognition}, 2022, pp. 9891--9901.

\bibitem{bhattacharyya2023ssl}
P.~Bhattacharyya, C.~Huang, and K.~Czarnecki, ``Ssl-lanes: Self-supervised learning for motion forecasting in autonomous driving,'' in \emph{Conference on Robot Learning}.\hskip 1em plus 0.5em minus 0.4em\relax PMLR, 2023, pp. 1793--1805.

\bibitem{chen2021multisiam}
K.~Chen, L.~Hong, H.~Xu, Z.~Li, and D.-Y. Yeung, ``Multisiam: Self-supervised multi-instance siamese representation learning for autonomous driving,'' in \emph{Proceedings of the IEEE/CVF International Conference on Computer Vision}, 2021, pp. 7546--7554.

\bibitem{zheng2021deep}
H.~Zheng, H.~Yong, and L.~Zhang, ``Deep convolutional dictionary learning for image denoising,'' in \emph{Proceedings of the IEEE/CVF conference on computer vision and pattern recognition}, 2021, pp. 630--641.

\bibitem{shao2020label}
S.~Shao, R.~Xu, W.~Liu, B.-D. Liu, and Y.-J. Wang, ``Label embedded dictionary learning for image classification,'' \emph{Neurocomputing}, vol. 385, pp. 122--131, 2020.

\bibitem{tang2020dictionary}
H.~Tang, H.~Liu, W.~Xiao, and N.~Sebe, ``When dictionary learning meets deep learning: Deep dictionary learning and coding network for image recognition with limited data,'' \emph{IEEE transactions on neural networks and learning systems}, vol.~32, no.~5, pp. 2129--2141, 2020.

\bibitem{zeng2023generative}
Z.~Zeng, R.~Zhu, Y.~Xia, H.~Zeng, and H.~Tong, ``Generative graph dictionary learning,'' in \emph{International Conference on Machine Learning}.\hskip 1em plus 0.5em minus 0.4em\relax PMLR, 2023, pp. 40\,749--40\,769.

\bibitem{wang2022repre}
L.~Wang, F.~Liang, Y.~Li, H.~Zhang, W.~Ouyang, and J.~Shao, ``Repre: Improving self-supervised vision transformer with reconstructive pre-training,'' \emph{arXiv preprint arXiv:2201.06857}, 2022.

\bibitem{liu2017sphereface}
W.~Liu, Y.~Wen, Z.~Yu, M.~Li, B.~Raj, and L.~Song, ``Sphereface: Deep hypersphere embedding for face recognition,'' in \emph{Proceedings of the IEEE conference on computer vision and pattern recognition}, 2017, pp. 212--220.

\bibitem{paszke2017automatic}
A.~Paszke, S.~Gross, S.~Chintala, G.~Chanan, E.~Yang, Z.~DeVito, Z.~Lin, A.~Desmaison, L.~Antiga, and A.~Lerer, ``Automatic differentiation in pytorch,'' 2017.

\bibitem{loshchilov2017decoupled}
I.~Loshchilov, ``Decoupled weight decay regularization,'' \emph{arXiv preprint arXiv:1711.05101}, 2017.

\bibitem{narayanan2018semi}
A.~Narayanan, Y.-T. Chen, and S.~Malla, ``Semi-supervised learning: Fusion of self-supervised, supervised learning, and multimodal cues for tactical driver behavior detection,'' \emph{arXiv preprint arXiv:1807.00864}, 2018.

\bibitem{tran2015learning}
D.~Tran, L.~Bourdev, R.~Fergus, L.~Torresani, and M.~Paluri, ``Learning spatiotemporal features with 3d convolutional networks,'' in \emph{Proceedings of the IEEE international conference on computer vision}, 2015, pp. 4489--4497.

\bibitem{wang2021oadtr}
X.~Wang, S.~Zhang, Z.~Qing, Y.~Shao, Z.~Zuo, C.~Gao, and N.~Sang, ``Oadtr: Online action detection with transformers,'' in \emph{Proceedings of the IEEE/CVF International Conference on Computer Vision}, 2021, pp. 7565--7575.

\end{thebibliography}
\end{document}